\documentclass[sigconf, natbib=false]{acmart}

\AtBeginDocument{%
  }

\setcopyright{acmlicensed}

\copyrightyear{2026}
\acmYear{2026}
\acmDOI{XXXXXXX.XXXXXXX}
\acmConference[SAC'26]{The 41st ACM/SIGAPP Symposium on Applied Computing}{March 23--27, 2026}{Thessaloniki, Greece}
\acmISBN{979-X-XXXX-XXXX-X/26/03}

\RequirePackage[
  datamodel=acmdatamodel,
  style=acmnumeric,
  ]{biblatex}

\usepackage{todonotes}
\usepackage{multirow}
\usepackage{graphicx}
\usepackage{algorithm}
\usepackage{algpseudocode}
\usepackage[table,xcdraw]{xcolor}

\begin{document}

\title{Adaptive Identification and Modeling of\\ Clinical Pathways with Process Mining}
\renewcommand{\shorttitle}{Adaptive Identification and Modeling of Clinical Pathways with Process Mining}

\author{Francesco Vitale}
\email{francesco.vitale@unina.it}
\affiliation{%
  \institution{University of Naples Federico II}
  \city{Naples}
  \country{Italy}
}

\author{Nicola Mazzocca}
\email{nicola.mazzocca@unina.it}
\affiliation{%
  \institution{University of Naples Federico II}
  \city{Naples}
  \country{Italy}
}

\renewcommand{\shortauthors}{F. Vitale and N. Mazzocca}

\begin{abstract}
Clinical pathways are specialized healthcare plans that model patient treatment procedures. They are developed to provide criteria-based progression and standardize patient treatment, thereby improving care, reducing resource use, and accelerating patient recovery. However, manual modeling of these pathways based on clinical guidelines and domain expertise is difficult and may not reflect the actual best practices for different variations or combinations of diseases. We propose a two-phase modeling method using process mining, which extends the knowledge base of clinical pathways by leveraging conformance checking diagnostics. In the first phase, historical data of a given disease is collected to capture treatment in the form of a process model. In the second phase, new data is compared against the reference model to verify conformance. Based on the conformance checking results, the knowledge base can be expanded with more specific models tailored to new variants or disease combinations. We demonstrate our approach using Synthea, a benchmark dataset simulating patient treatments for SARS-CoV-2 infections with varying COVID-19 complications. The results show that our method enables expanding the knowledge base of clinical pathways with sufficient precision, peaking to 95.62\% AUC while maintaining an arc-degree simplicity of 67.11\%.
\end{abstract}



\maketitle

\section{Introduction}
\label{sec:intro}

Emerging infectious diseases continue to pose significant challenges worldwide, influenced by factors such as globalization, urbanization, and climate change \cite{baker2022infectiousdiseaseglobalchange}. For example, the recent COVID-19 pandemic has had tremendous effects on several key socio-economic sectors due to its high transmissibility and health complications. Among the most concerning aspects of the pandemic was the high variability in the clinical course of the infected patients \cite{flisiak2023variability}. In fact, the incubation period of the virus is variable and it can be difficult to identify early symptoms or asymptomatic cases. Clinical outcomes also vary widely, from mild cold-like symptoms to severe pneumonia and respiratory syndromes \cite{whiteson2023pulmonary}. Moreover, the pandemic unfolded in multiple waves of different SARS-CoV-2 variants such as Delta and Omicron, which differed in transmissibility, disease severity, and immune escape \cite{zabidi2023sarscov2variants}.

In view of the significant challenges posed by the variability of not only the COVID-19 disease, but also others such as the Ebola virus disease and HIV/AIDS, the development of clinical pathways, which are specialized healthcare plans that model patient treatment procedures, can address the different courses, complications and outcomes of these diseases \cite{croce2016hivclinicalpathway,rojek2017ebolaoutbreak,xu2020covid19clinicalpathway}. More specifically, a recent study \cite{lawal2016clinicalpathwaysystematicreview} provided a practical operational definition of clinical pathways, involving structured multidisciplinary plan of care, translation of guidelines or evidence into local structures, description of steps in a course of treatment, criteria-based progression, and standardization of the care process for a specific clinical problem and a specific population. The effort of developing effective clinical pathways produces short- and long-term benefits, such as improved care and quality of health, reduction of resource use, detailed guidance for the clinical treatment, and acceleration of patient recovery \cite{rotter2025clinical}. 

Clinical pathways can be developed by combining clinical guidelines and domain experience/expertise. For example, Croce et al. \cite{croce2016hivclinicalpathway} assessed the impact of a clinical pathway implementation, evaluating effectiveness, efficiency, and cost reduction of treating HIV-infected adults. They defined the clinical pathway by engaging with a multi-disciplinary team of economists, clinicians and pharmacists, and interviewing the most recent national and international HIV guidelines. Another example is that of Xu et al. \cite{xu2020covid19clinicalpathway}, who developed a clinical pathway for early diagnosis of COVID-19 by reviewing the evidence of multiple reports in the literature and the clinical protocols of the national health commission of China. Thus, the clinical pathway was a result of the combination of clinical experience and guidelines.

Although clinical pathways are an inestimable tool, their development is rather challenging without the support of data-driven methods. In this context, process mining has seen widespread use in healthcare, as it can be adopted due to its ability to handle event data and produce process models, which can provide interpretable insights on clinical pathways bottlenecks and inefficiencies \cite{aversano2025pmhealthcareapplications}. However, process models lose their usefulness when they become too complex or do not adequately capture the intended behavior. Many works \cite{litchfield2018pmlongtermconditionsdescription, munoz2022process, cuendet2022covid19managementpmanalysis, bossonrieutort2025healthcaretrajectoriesprocessmining} in healthcare often report the ``spaghetti" effect, which has been long documented and addressed in theoretical process mining works \cite{aalst2016pm, vidgof2020spaghetti}. 

The existing applications of process mining for healthcare attempt to address the aforementioned challenges through extensive data preprocessing. However, the inherent complexity of the healthcare processes cannot be solely addressed by cleaning the data; a systematic approach balancing well-known quality criteria independent of data quality is needed. Additionally, the existing works do not deal with the possibility of concept drift, which involves changes in the activity executions of target processes \cite{adams2023explainableconceptdrift}. In light of these challenges, we propose process mining-based method whose novelties lie in:
\begin{itemize}
    \item the ability to model clinical pathways adaptively to facilitate the identification of different disorders;
    \item keep the complexity of the identified clinical pathways manageable to facilitate the models' interpretability.
\end{itemize}

The proposal achieves these goals by combining well-known process mining algorithms with machine learning-based post-processing. The method follows two phases: the first phase collects historical event logs from existing clinical courses where patients were treated for specific diseases. These logs are filtered and subsequently captured into a prescriptive process model. The second phase involves monitoring the treatment processes and verifying the conformance against the prescriptive process model(s). If the degree of conformance is unsatisfying, new models are mined to keep a meaningful knowledge base.

We experiment with our method using the Synthea COVID-19 dataset \cite{walonoski2020synthea}, a well-known healthcare benchmark for practitioners and researchers that includes the synthetic generation of COVID-19 patients' clinical courses affected by different health complications. The dataset includes the clinical courses of several patients under different COVID-19 complications. Results show that our method enables expanding the knowledge base of clinical pathways with sufficient precision, striking a balance between specificity and generalization. 


The rest of the paper is organized as follows. Section \ref{sec:related_work} reviews related work on process mining in healthcare and its use for clinical pathway modeling; Section \ref{sec:method} describes the proposed process mining-based method for adaptive identification and modeling of clinical pathways; Section \ref{sec:evaluation} evaluates our method's capabilities with the Synthea COVID-19 dataset; and Section \ref{sec:conclusions} concludes by summarizing the findings and reviewing future work.

\section{Related work}
\label{sec:related_work}

\subsection{Process mining in healthcare}
Process mining ``brings together traditional model-based process analysis and data-centric analysis techniques'' by allowing data-driven discovery of process models (process discovery) and verifying the conformance of new data against such models (conformance checking) \cite{aalst2016pm}. This is particularly valuable in complex environments such as healthcare, where gaining a bird’s-eye view of actual processes is essential and deviations from guidelines and standards are both common and inevitable.

The application of process mining in healthcare spans a wide range of application contexts, including clinical pathways, and healthcare specialties, including oncology, cardiovascular diseases, respiratory diseases, etc. \cite{erdogan2018pmhealthcaresystematicmappingstudy, aversano2025pmhealthcareapplications}. However, there are several challenges associated with the application of process mining, mainly due to the complex, limited, and highly heterogeneous nature of the data collected in healthcare information systems \cite{munoz2022process,naeimaei2023clinical}.  

While the above-mentioned data challenges are worthwhile mentioning, in this paper, we focus on addressing the inherent complexity of the healthcare processes. In fact, while data issues influence the quality and applicability of process mining, the underlying process can be naturally prone to so-called \textit{concept drifts}, which are changes in the executed process independent of data collection \cite{adams2023explainableconceptdrift}.

\begin{figure*}[!t]
\centering
\includegraphics[width=\textwidth]{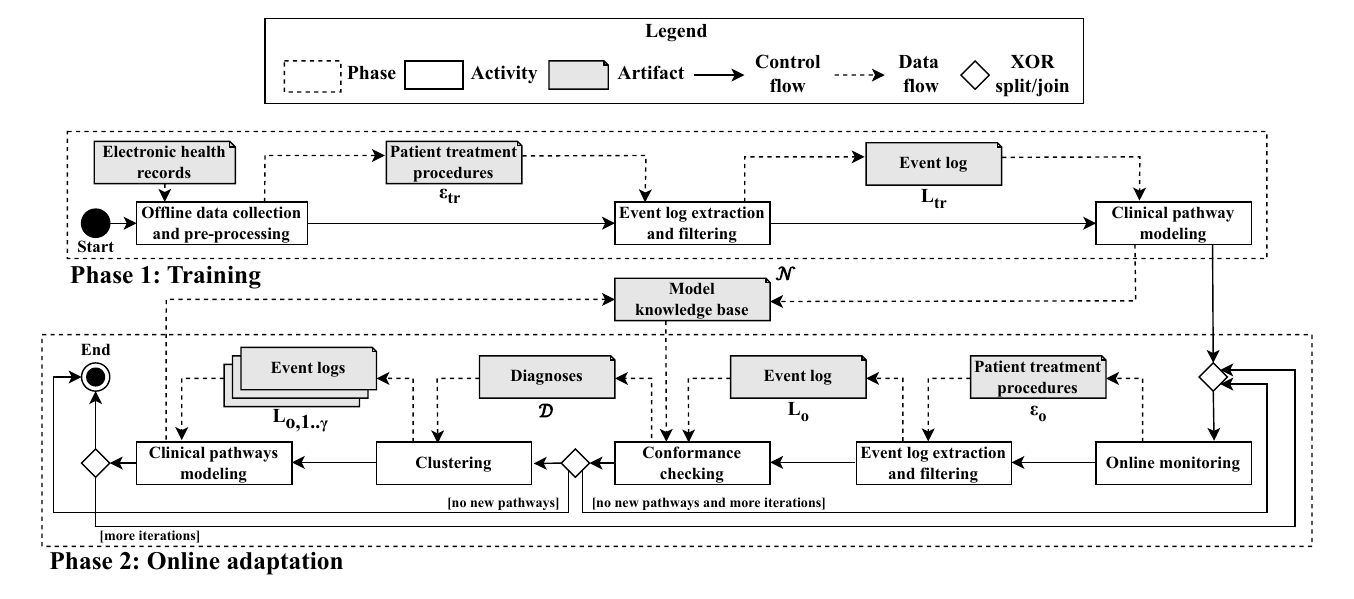}
\caption{The proposed two-phase method for adaptive identification and modeling of clinical pathways in healthcare.}
\label{fig:method}
\end{figure*}

\subsection{Process mining for clinical pathway analysis}
Among the promising use cases for process mining in healthcare, the ability to analyze clinical pathways of healthcare patients has seen increasing interest by researchers and use by practitioners.

Phan et al. \cite{phan2019pmcpanalysis} investigated the use of process mining to model clinical pathways for incisional hernia treatment, emphasizing the value of analyzing procedures and outcomes to reduce complications and lower healthcare costs. Similarly, Cuendet et al. \cite{cuendet2022covid19managementpmanalysis} applied process mining to examine COVID-19 treatment management across patients with varying comorbidities, particularly those with active cancer during two pandemic waves, demonstrating how such methods can deepen understanding of resource utilization in complex care settings. Extending this line of inquiry, Yang et al. \cite{yang2023acutepancreatitisclinicalpathwaysprocessmining} explored clinical pathways in acute pancreatitis and underscored the difficulty of interpreting convoluted processes that arise from highly complex electronic medical record data.

Other studies further highlight both the promise and challenges of applying process mining in healthcare. Li et al. \cite{li2024strokerehabilitationclinicalpathway} examined cerebral stroke rehabilitation pathways, showing that process mining can reveal real-world treatment dynamics while still producing models that may be difficult to read. Beyel et al. \cite{beyel2024heartfailureprocessawareanalysis} focused on heart failure treatment trajectories, evaluating the ability of mined models to uncover causes of cardiovascular complications and to predict future events. Finally, Bosson-Rieutort et al. \cite{bossonrieutort2025healthcaretrajectoriesprocessmining} analyzed healthcare trajectories during patients’ final year of life, offering insights into care planning for aging populations but also highlighting substantial limitations, including spaghetti-like process representations caused by noisy and heterogeneous healthcare data.

Collectively, these studies highlight the growing adoption of process mining in clinical pathway analysis. The technique offers interpretable representations of patient care pathways and the potential to predict outcomes based on current treatments. Nevertheless, substantial challenges remain. Mining simple yet meaningful process models is difficult due to both algorithmic representation biases and the noisy nature of healthcare data. In this study, we propose an adaptive approach for the online and autonomous discovery of simplified process models capable of accurately identifying distinct pathways associated with different disease complications.

\section{Method}
\label{sec:method}

The proposed method aims to adaptively model clinical pathways as new, unseen treatments are filed in electronic health records. To identify new clinical pathways and include them in a progressively richer knowledge base, the proposal is organized into two phases: the training and online adaptation phases. The main novelty of the work lies in the second phase, which adaptively learns new clinical pathways from online data on different patient treatment procedures.

The method is shown in Figure \ref{fig:method}. The flow diagram distributes two sets of steps (solid boxes) between the training and online adaptation phases (dashed boxes). The control flow (solid lines and diamond nodes) logically connects the flow of steps, while the data flow (dashed lines) reports the inputs and outputs of the steps. The two phases are detailed in the following.

\subsection{Phase 1: Training}
The goal of this phase is to mine a baseline clinical pathway to use as a reference for identifying conforming and deviating treatments in the online adaptation phase. To this aim, the phase starts with \textbf{offline data collection and pre-processing} from \textit{electronic health records}. This activity is crucial since these records may capture unstructured data about the patients' treatments, whereas the application of process mining algorithms requires well-structured data, where events are easily recognizable. An event $e$ is defined as a $k$-tuple of attributes $(at_1,at_2,\dots,at_k)$, where each attribute can either be categorical or numerical. Since process mining algorithms are primarily concerned with processes' control flow \cite{aalst2016pm}, the attributes must include: the timestamp, an activity, and a unique event identifier. Let us denote $\mathcal{E}_{tr}=\{e_1,e_2,\dots,e_{K_{tr}}\}$ the set of $K_{tr}$ \textit{patient treatment procedures}. We further organize these procedures into a set of cases $\mathcal{C}_{tr}=\{C_1,C_2,\dots,C_\psi\}$, where a case $C\in\mathcal{C}_{tr}$ groups a disjunct set of events, i.e., $\nexists\, e\in\mathcal{E}_{tr}:e\in C_i\land e\in C_j, \forall i\neq j$. Each case represents a given treatment flow of a patient. In particular, given case $C=\{e_1,e_2,\dots,e_{|C|}\}\in \mathcal{C}_{tr}$ and $a_{e_i}$ the activity associated with the $i$-th event $e\in C$, the sequence of activities $\sigma=\langle a_{e_1}, a_{e_2},\dots,a_{e_{|C|}}\rangle$ associated with the events is the treatment flow of a patient.

The patient treatment procedures are handled by the \textbf{event log extraction and filtering} step. In this work, we will only focus on the control flow perspective and define an event log as follows. Given $\mathcal{A}$ the set of activities associated with patient treatment procedures, $\mathcal{A}^*$ the set of all finite sequences of such procedures, and $\mathbb{B}(\mathcal{A}^*)$ the set of bags over $\mathcal{A}^*$, the event log $L_{tr}\in \mathbb{B}(\mathcal{A}^*)$ associated with $\mathcal{C}_{tr}$ is a multi-set of finite sequences over $\mathcal{A}$. $L_{tr}$ may contain a multitude of activities, of which some may constitute noise. The effect of noise on process mining algorithms can lead to very convoluted models, contributing to the ``spaghetti'' effect of process models often seen in practice. To reduce the impact of noise on the subsequent modeling activity, we include variant-based filtering, which can leave out infrequent variants and generate a filtered, higher-quality event log. A practical approach can be to order the variants of an event log based on their frequency of occurrence and choose the top-$k$ variants.

\begin{figure*}[!t]
\centering
\includegraphics[width=\textwidth]{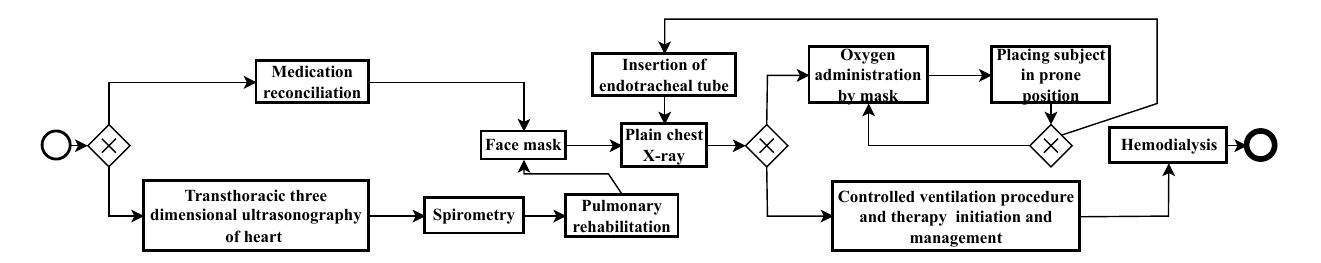}
\caption{A BPMN model describing different procedures in a clinical pathway mined through our method.}
\label{fig:bpmn}
\end{figure*}
Finally, the filtered event log is used for \textbf{clinical pathway modeling}, which generates a process model $N$ that represents the clinical pathway. This can be performed through process discovery, which captures control-flow relationships between the activities of $L_{tr}$. The structure of $N$ depends on the process discovery algorithm bias and the process structure captured by the activities of $L_{tr}$. Additionally, different types of formalisms can capture the model, such as the Business Process Modeling Notation (BPMN) \cite{pufahl2022bpmn} and the Petri net \cite{mahulea2018modular}. Figure \ref{fig:bpmn} shows an example BPMN model from our experimentation, which is usually preferred over Petri nets to model business processes due to its heightened interpretability. $N$ is stored and used as the baseline in the model knowledge base $\mathcal{N}$ for further monitoring patient treatment procedures in the subsequent online phase.

\subsection{Phase 2: Online adaptation}
This phase unfolds in different iterations. For each iteration, the \textbf{online monitoring} of patient treatment procedures $\mathcal{E}_{o}$ is performed to monitor the treatment flow implemented by clinicians to treat a given disease. These procedures undergo the same event log extraction and filtering step as done in the training phase to ensure consistent processing of the data, resulting in the event log $L_o$. 

\begin{table}[!t]
\centering
\caption{Alignment-based conformance checking diagnoses obtained from replaying the traces $\sigma_1\dots\sigma_{|L_o|}$ of event log $L_o$ against a reference Petri net $N$.}
\label{cc_diagnoses}
\begin{tabular}{llllll}
\hline
\textbf{Trace}                & $\boldsymbol{a_1}$       & $\boldsymbol{a_2}$       & $\cdots$ & $\boldsymbol{a_{|\mathcal{A}|}}$       & $\boldsymbol{F_{\sigma}}$ \\ \hline
$\boldsymbol{\sigma_1}$       & $u_{a_1,\sigma_1}$       & $u_{a_2,\sigma_1}$       & $\cdots$ & $u_{a_{|\mathcal{A}|},\sigma_1}$       & $F_{\sigma_1}$            \\
$\boldsymbol{\sigma_2}$       & $u_{a_1,\sigma_2}$       & $u_{a_2,\sigma_2}$       & $\cdots$ & $u_{a_{|\mathcal{A}|},\sigma_2}$       & $F_{\sigma_2}$            \\
$\cdots$                      & $\cdots$                 & $\cdots$                 & $\cdots$ & $\cdots$                               & $\cdots$                  \\
$\boldsymbol{\sigma_{|L_o|}}$ & $u_{a_1,\sigma_{|L_o|}}$ & $u_{a_2,\sigma_{|L_o|}}$ & $\cdots$ & $u_{a_{|\mathcal{A}|},\sigma_{|L_o|}}$ & $F_{\sigma_{|L_o|}}$      \\ \hline
\end{tabular}
\end{table}
Next, \textbf{conformance checking} is performed to compare $L_o$ with the models of the knowledge base $\mathcal{N}$. Given a reference model $N\in\mathcal{N}$, the class of process mining conformance checking algorithms involves evaluating how close the traces of $L_o$ align with $N$. The output of such an evaluation for each trace $\sigma\in L_o$ is the so-called fitness $F_\sigma\in[0,1]$, which quantifies the degree of conformance between $N_i$ and $\sigma$ as a real number between 0 and 1, where the former indicates complete non-conformance, whereas the latter indicates full conformance. Conformance checking allows recording other information apart from fitness, such as non-existing/duplicated/wrongly-ordered activities. These non-conformances are referred to as conformance-checking \textit{diagnoses} $\mathcal{D}\in\mathbb{N}^{|L_o|\times|\mathcal{A}|}$ \cite{vitale2025pmdt,vitale2025cfad}, shown in Table \ref{cc_diagnoses}. Each element $u_{a_i,\sigma}\in\mathcal{D}$ is an integer that evaluates the mismatches linked to activity $a_i\in\mathcal{A}$ between $\sigma\in L_o$ and $N\in\mathcal{N}$.
Our method builds the diagnoses by considering the best model in the knowledge base $\mathcal{N}$. Algorithm \ref{best_petri_net_algorithm} describes the steps to build the diagnoses. For each trace $\sigma\in L_o$, the best diagnosis is selected among those calculated for each $N\in\mathcal{N}$.
\begin{algorithm}[!t]
\caption{Diagnoses computation}
\label{best_petri_net_algorithm}
\begin{algorithmic}[1]
\State \textbf{Input:} $L_o = \{\sigma_{1},\dots,\sigma_{|L_o|}\}$, Petri nets $\mathcal{N}$
\State \textbf{Output:} $\mathcal{D}\in\mathbb{N}^{|L_o|\times|\mathcal{A}|}$

\State $\mathcal{D}$ $\gets \{\}$
\For{$\sigma\in L_o$}
    \State best\_fitness $\gets -\infty$
    \State best\_diagnosis $\gets \{\}$
    \For{$N \in \mathcal{N}$}
        \State $\text{diagnosis} \gets \text{conformance\_checking}(\sigma), N)$
        \If{\text{diagnosis}($F_\sigma$) > best\_fitness}
            \State best\_fitness $\gets \text{diagnosis}(F_\sigma)$
            \State best\_diagnosis $\gets \text{diagnosis}$
        \EndIf
    \EndFor
    \State $\mathcal{D}\gets \mathcal{D} \cup \{\text{best\_diagnosis}\}$
\EndFor
\State \Return $\mathcal{D}$
\end{algorithmic}
\end{algorithm}

To evaluate whether a new disease is being treated differently than the pathways stored in $\mathcal{N}$, a threshold $th$ on the fitness value can be set. Specifically, given $F_{1,\dots,|\mathcal{N}|}$ the set of fitness values obtained by comparing $L_o$ with each $N\in\mathcal{N}$, if there exists no $F\in F_{1,\dots,|\mathcal{N}|}$ such that $F\geq th$, then $L_o$ contains new unseen clinical pathways. 

Finally, to capture the new clinical pathways, \textbf{clustering} groups the diagnoses of non-conformant traces so that new clinical pathways can be mined and included in the model knowledge base $\mathcal{N}$. Specifically, we select the diagnoses obtained with the most conforming Petri net and perform clustering with these diagnoses. Given $\gamma$ clusters, the event log $L_o$ is split into $\gamma$ sublogs $L_{o,1\dots,\gamma}$. Each sublog is handled independently by process discovery to generate new process models and expand the model knowledge base.

\section{Evaluation}
\label{sec:evaluation}

In this section, we provide an overview of the Synthea dataset, the benchmark used to evaluate the performance of our proposal. Then, the experimental setup is detailed in regards to the goals and metrics used to assess our results. Finally, the classification results of our method are shown and discussed.

\subsection{Dataset and experimental setup}
The Synthea dataset\footnote{https://synthea.mitre.org/downloads} records synthetic electronic health records collected from the realistic simulation of COVID-19 patient treatments \cite{walonoski2020synthea}. The data collect clinical courses, outcomes, and healthcare utilization of patients with COVID-19. They include over 124,000 synthetic patients, of whom about 88,000 were infected. Furthermore, the dataset captures key epidemiological features such as hospitalization ($\approx$20\%), mortality ($\approx$4\%), ICU and inpatient length of stay, as well as detailed daily vitals, laboratory values, symptom trajectories, and complications like ARDS or sepsis.

\begin{table}[!t]
\centering
\caption{Properties of the event logs extracted from the Synthea dataset.}
\label{dataset_stats}
\begin{tabular}{llll}
\hline
\textbf{Event log} & \textbf{\# Activites} & \textbf{\# Traces} & \textbf{Trace length} \\ \hline
$L_{PE}$           & 1521                  & 104                & 14$\pm$13             \\
$L_{ARD}$          & 8053                  & 234                & 34$\pm$16             \\
$L_{SST}$          & 1231                  & 132                & 9$\pm$12              \\
$L_{AVP}$          & 4866                  & 514                & 9$\pm$14              \\
$L_{AB}$           & 5051                  & 463                & 10$\pm$14             \\
$L_{MNB}$          & 2606                  & 158                & 16$\pm$14             \\ \hline
\end{tabular}
\end{table}
\begin{table*}[!t]
\centering
\resizebox{\textwidth}{!}{\begin{tabular}{lllllllllllllllll}
\hline
\multicolumn{2}{l}{\textbf{Method}} & \multicolumn{3}{l}{\textbf{Iteration \#1}}                        &  & \multicolumn{3}{l}{\textbf{Iteration \#2}}                        & \textbf{} & \multicolumn{3}{l}{\textbf{Iteration \#3}}                        & \textbf{} & \multicolumn{3}{l}{\textbf{Iteration \#4}}                        \\ \cline{3-5} \cline{7-9} \cline{11-13} \cline{15-17} 
\multicolumn{2}{l}{}                & \textbf{AUC (\%)}          & \textbf{S (\%)} & \textbf{\# Models} &  & \textbf{AUC (\%)}          & \textbf{S (\%)} & \textbf{\# Models} & \textbf{} & \textbf{AUC (\%)}          & \textbf{S (\%)} & \textbf{\# Models} & \textbf{} & \textbf{AUC (\%)}          & \textbf{S (\%)} & \textbf{\# Models} \\ \hline
ILP               & DBScan          & $90.04_{1.73}$             & $32.50_{0.26}$  & $3_{0}$            &  & $89.40_{0.62}$             & $69.01_{5.69}$  & $9_{2}$            &           & $93.90_{0.26}$             & $70.08_{8.95}$  & $15_{3}$           &           & $\underline{95.62_{0.11}}$ & $67.11_{7.67}$  & $19_{3}$           \\
\textbf{}         & OPTICS          & $90.11_{1.87}$             & $20.16_{0.32}$  & $4_{0}$            &  & $\underline{89.62_{0.39}}$ & $43.47_{4.80}$  & $8_{1}$            &           & $92.91_{0.40}$             & $52.33_{0.05}$  & $13_{1}$           &           & $94.62_{0.39}$             & $50.58_{0.98}$  & $16_{1}$           \\
                  & DPGMM           & $89.98_{1.91}$             & $32.37_{0.10}$  & $10_{1}$           &  & $89.11_{0.77}$             & $44.81_{0.80}$  & $16_{2}$           &           & $\underline{93.82_{0.10}}$ & $45.47_{0.22}$  & $20_{2}$           &           & $95.41_{0.17}$             & $44.07_{1.64}$  & $22_{3}$           \\
\textbf{}         & Baseline        & $87.70_{2.74}$             & $9.27_{0.58}$   & $1_{0}$            &  & $77.71_{16.92}$            & $8.19_{0.04}$   & $1_{0}$            &           & $67.04_{3.41}$             & $15.79_{0.00}$  & $1_{0}$            &           & $77.21_{3.46}$             & $17.67_{1.65}$  & $1_{0}$            \\ \hline
IM                & DBScan          & $80.92_{5.86}$             & $79.23_{1.60}$  & $3_{0}$            &  & $82.42_{4.21}$             & $90.76_{1.44}$  & $9_{0}$            &           & $90.97_{2.03}$             & $90.02_{0.48}$  & $15_{0}$           &           & $93.64_{1.60}$             & $90.90_{0.68}$  & $21_{1}$           \\
\textbf{}         & OPTICS          & $80.58_{4.29}$             & $78.75_{1.66}$  & $4_{0}$            &  & $81.81_{3.25}$             & $87.61_{0.67}$  & $9_{0}$            &           & $89.86_{2.70}$             & $89.06_{0.30}$  & $14_{0}$           &           & $92.37_{2.30}$             & $88.40_{0.96}$  & $17_{1}$           \\
                  & DPGMM           & $81.36_{5.24}$             & $81.79_{1.51}$  & $8_{0}$            &  & $82.89_{4.60}$             & $86.02_{1.05}$  & $15_{1}$           &           & $89.79_{3.66}$             & $85.14_{1.40}$  & $17_{1}$           &           & $92.07_{3.14}$             & $84.32_{1.72}$  & $20_{1}$           \\
\textbf{}         & Baseline        & $85.58_{1.83}$             & $85.81_{3.43}$  & $1_{0}$            &  & $84.06_{1.90}$             & $78.82_{3.27}$  & $1_{0}$            &           & $84.88_{3.72}$             & $72.46_{0.73}$  & $1_{0}$            &           & $80.31_{7.16}$             & $70.20_{1.74}$  & $1_{0}$            \\ \hline
HM                & DBScan          & $87.73_{2.25}$             & $74.08_{0.85}$  & $2_{0}$            &  & $88.08_{0.92}$             & $86.20_{1.93}$  & $7_{1}$            &           & $92.82_{1.01}$             & $86.79_{0.99}$  & $13_{0}$           &           & $94.21_{0.04}$             & $86.93_{0.99}$  & $19_{1}$           \\
                  & OPTICS          & $88.20_{1.92}$             & $77.17_{3.04}$  & $4_{0}$            &  & $88.15_{0.46}$             & $86.35_{2.45}$  & $8_{0}$            &           & $91.78_{0.10}$             & $86.17_{1.22}$  & $12_{0}$           &           & $93.19_{0.67}$             & $85.90_{0.62}$  & $16_{1}$           \\
                  & DPGMM           & $\underline{89.71_{3.73}}$ & $87.66_{0.61}$  & $9_{2}$            &  & $89.25_{2.33}$             & $90.91_{0.07}$  & $15_{2}$           &           & $93.67_{0.73}$             & $87.82_{0.23}$  & $18_{2}$           &           & $94.82_{0.00}$             & $86.52_{0.12}$  & $20_{2}$           \\
                  & Baseline        & $77.11_{9.98}$             & $66.46_{0.50}$  & $1_{0}$            &  & $81.55_{5.90}$             & $64.26_{0.46}$  & $1_{0}$            &           & $86.58_{1.42}$             & $61.33_{0.41}$  & $1_{0}$            &           & $79.68_{5.07}$             & $58.99_{0.27}$  & $1_{0}$            \\ \hline
\end{tabular}}
\caption{The Area Under the Receiving Operating Curve (AUC) percentage, Simplicity (S) percentage and number of models for each process discovery-clustering combination of our method, together with the baseline process discovery approaches, per iteration. The underlined figures highlight the best AUC values per iteration.}
\label{tab:results}
\end{table*}
To evaluate the proposed method and its ability to adaptively update the model knowledge base with new clinical pathways, we extracted five sets of event logs from the database. We first selected all patients diagnosed with COVID-19. Then, we identified patients with disorders clinically associated with COVID-19, namely: pulmonary emphysema (PE), acute respiratory distress syndrome (ARD), streptococcal sore throat (SST), acute viral pharyngitis (AVP), acute bronchitis (AB) and malignant neoplasm of the breast (MNB). Although these diseases are associated with COVID-19, their evolution and clinical pathway are expected to be different due to different severity and complications. For each patient, treatment procedures were recorded and filtered by retaining the $20$ most frequent variants, in order to remove noise and rare outliers. Each dataset was then converted into an event log, denoted as $L_{PE}$, $L_{ARD}$, $L_{SST}$, $L_{AVP}$, $L_{AB}$, and $L_{MNB}$, respectively. The statistics of each event log are reported in Table \ref{dataset_stats}.

\begin{figure*}[!t]
\centering
\includegraphics[width=\textwidth]{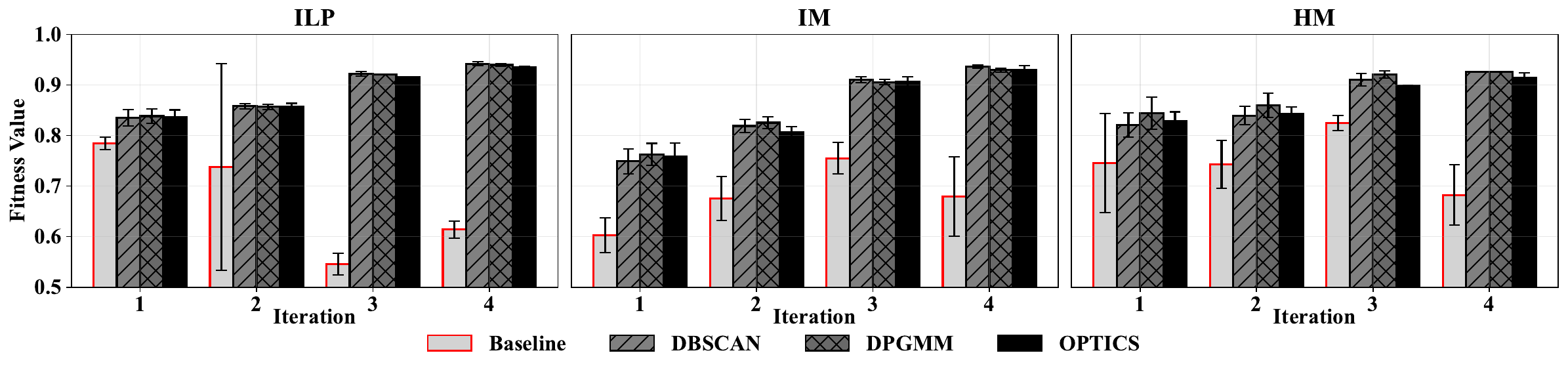}
\caption{The fitness values per process discovery algorithm and clustering technique for each iteration, in comparison with the baseline.}
\label{fig:fitness}
\end{figure*}
\begin{figure*}[!t]
\centering
\includegraphics[width=\textwidth]{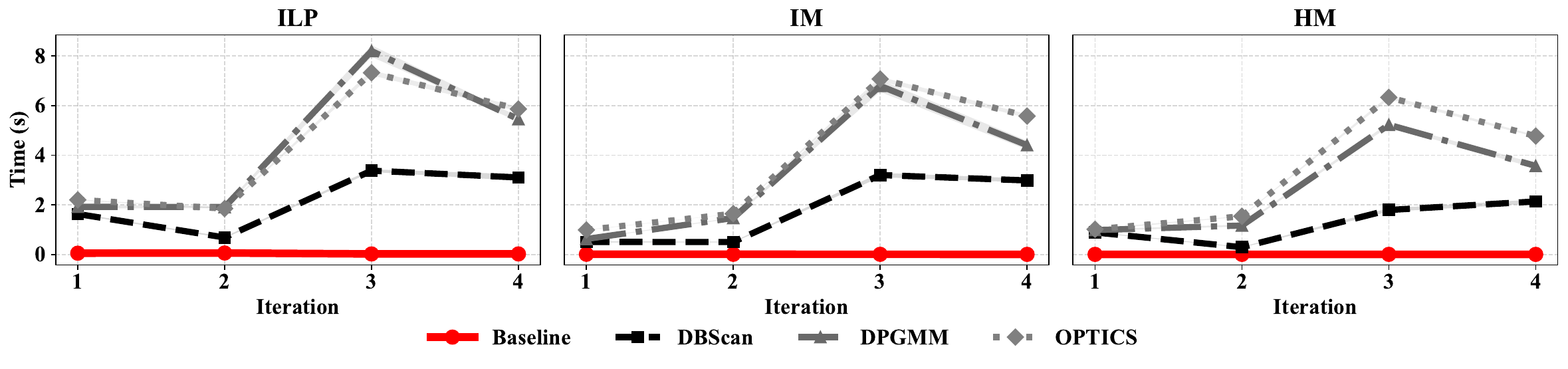}
\caption{The time required for batch processing of the event logs during online adaptation at each iteration for all the methods.}
\label{TIMING_ANALYSIS}
\end{figure*}

The event logs were used in two stages. First, $L_{tr}$ was built using 75\% of $L_{PE}$. Then, in phase 2, the online adaptation phase was simulated across four iterations, one for each of the remaining event logs. In the $i$-th iteration, let $L_{o,i}$ denote the log. For each iteration, 75\% of $L_{o,i}$ was used for online adaptation, while the remaining 25\%, denoted $L_{o,i,tst}$, together with the 25\% of each of the previous iterations' event logs, was reserved for testing. We considered a total of 4 iterations, in which the event logs $L_{ARD}$, $L_{SST}$, $L_{AVP}$ and $L_{AB}$ are split as reported above. We label the traces of these sets as ``negative'' traces, as these should be treated and identified as legitimate clinical pathways. On the other hand, the traces of $L_{MNB}$ are always included in the test set, and labeled as ``positive'' traces. We did so to simulate the case where a type of clinical pathway is not legitimate and should be flagged as anomalous.

The metrics used to evaluate the quality of modeling and adaptation are the arc-degree simplicity (S), the total number of models, and the Area Under the ROC Curve (AUC). The arc-degree simplicity allows evaluating the complexity of the process models based on the incoming and outgoing arcs of each model element \cite{vazquez2015prodigen}, and is a normalized measure between 0 and 1. S values close to 0 indicate a high arc degree across the process model's nodes, whereas values close to 1 indicate arc degrees approximately equal to 2. The total number of models provides information about the fragmentation of the clinical pathways into different models. The AUC measures the ability to classify the traces of the test event logs of each iteration using the fitness measure \cite{bradley1997use}. We set 9 configurations for our method, involving 3 process discovery algorithms, namely ILP \cite{vanzelst2018ilp}, IM \cite{leemans2013discovering}, and HM \cite{weijters2011fhm}, and 3 clustering algorithms, namely DBScan \cite{ester1996density}, OPTICS \cite{ankerst1999optics}, and DPGMM \cite{gorur2010dirichlet}. We have repeated the experiment for each combination of process discovery and clustering 3 times to account for randomicity in the event log splitting.

For each process discovery algorithm we have also included the baseline case, for which there is no adaptation. This influences the split we did earlier for each iteration of the online phase. In this case, 75\% of traces of $L_{o,i}$ used for training are joined with the 75\% of traces of iteration $i-1$, building a single process model that captures information of all the traces up until the $i$-th iteration.

The software is implemented using Python and uses several process mining and machine learning packages, including \texttt{pm4py} and \texttt{scikit-learn}. The code is available online on GitHub\footnote{https://github.com/francescovitale/pm\_cp\_adaptation}.

\subsection{Results}
Table \ref{tab:results} reports the results in terms of AUC and S percentages, as well as the number of models obtained in each iteration. The method is not only able to keep high AUC percentages across all iterations, but it also highlights an increasing trend, which means that adaptation is increasingly effective at fragmenting the diagnoses to build better process models. For example, considering the HM combined with the DPGMM, AUC goes from 89.71\% up to 94.82\% while maintaining a high average simplicity of the process models, from 87.66\% to 86.52\%. The downside of this trend is the increasing number of models, which goes from 9 in iteration \#1 to 20 in iteration \#4. 

Remarkably, the baseline is unable to maintain the same performance. Generally, the AUC tends to be lower than the other methods, apart from the case of IM, for which the baseline is better in iterations \#1 and \#2. However, the actual problem with the baseline is the lower S values, especially for the ILP algorithm. In fact, for this algorithm, at iteration \#4, S is equal to 17.67\% (ILP baseline), 70.20\% (IM baseline), and 58.99\% (HM baseline), whereas the proposed method achieves as high as 67.11\% (ILP with DBScan), 90.90\% (IM with DBScan), and 86.93\% (HM with DBScan). 

The interpretation of these results can be found in Figure \ref{fig:fitness}. These bar plots show how the mean fitness values of the negative traces vary at each iteration for each process discovery-clustering/baseline combination. The baseline fitness values are generally lower than those achieved by the method combining the clustering approach, showing better adaptation to new conditions. This improved adaptation occurs without sacrificing simplicity, which remains high across all iterations.

\begin{table}[!t]
\centering
\caption{Two-way ANOVA and group means with 95\% Confidence Intervals (CI) across the four iterations per process discovery algorithm. The grey rows outline the best process discovery-clustering combinations.}
\label{ANOVA_RESULTS}
\begin{tabular}{llllll}
\hline
\multicolumn{2}{l}{Method} & \multicolumn{2}{l}{\textbf{AUC (\%)}}                               & \multicolumn{2}{l}{\textbf{Simplicity (\%)}}                        \\ \cline{3-6} 
      & \textbf{}          & \textbf{Mean}                & \textbf{95\% CI}                     & \textbf{Mean}                & \textbf{95\% CI}                     \\ \hline
ILP   & DBScan             & \cellcolor[HTML]{E8E8E8}91.8 & \cellcolor[HTML]{E8E8E8}[89.8, 93.7] & \cellcolor[HTML]{E8E8E8}55.8 & \cellcolor[HTML]{E8E8E8}[46.9, 64.7] \\
      & OPTICS             & 91.3                         & [89.7, 92.8]                         & 39.2                         & [31.6, 46.7]                         \\
      & DPGMM              & 91.6                         & [89.6, 93.6]                         & 41.4                         & [38.2, 44.6]                         \\
      & Baseline           & 77.4                         & [69.8, 85.1]                         & 12.7                         & [10.0, 15.5]                         \\ \cline{2-6} 
      & \textit{p-values}  & \multicolumn{2}{l}{\textbf{Clustering: $<0.001$}}                   & \multicolumn{2}{l}{\textbf{Clustering: $<0.001$}}                   \\
      & \textit{}          & \multicolumn{2}{l}{\textbf{Iteration: $0.209$}}                     & \multicolumn{2}{l}{\textbf{Iteration: $<0.001$}}                    \\ \hline
IM    & DBScan             & \cellcolor[HTML]{E8E8E8}84.3 & \cellcolor[HTML]{E8E8E8}[79.5, 89.2] & \cellcolor[HTML]{E8E8E8}87.6 & \cellcolor[HTML]{E8E8E8}[83.9, 91.3] \\
      & OPTICS             & 83.9                         & [79.6, 88.1]                         & 84.7                         & [81.9, 87.4]                         \\
      & DPGMM              & 83.3                         & [79.2, 87.4]                         & 83.3                         & [82.2, 84.5]                         \\
      & Baseline           & 83.7                         & [80.6, 86.8]                         & 76.8                         & [72.4, 81.2]                         \\ \cline{2-6} 
      & \textit{p-values}  & \multicolumn{2}{l}{\textbf{Clustering: $0.935$}}                    & \multicolumn{2}{l}{\textbf{Clustering: $<0.001$}}                   \\
      & \textit{}          & \multicolumn{2}{l}{\textbf{Iteration: $<0.001$}}                    & \multicolumn{2}{l}{\textbf{Iteration: $<0.001$}}                    \\ \hline
HM    & DBScan             & 90.0                         & [87.8, 92.2]                         & 83.8                         & [80.2, 87.5]                         \\
      & OPTICS             & 90.0                         & [88.2, 91.9]                         & 83.7                         & [82.0, 85.4]                         \\
      & DPGMM              & \cellcolor[HTML]{E8E8E8}90.5 & \cellcolor[HTML]{E8E8E8}[88.4, 92.7] & \cellcolor[HTML]{E8E8E8}88.3 & \cellcolor[HTML]{E8E8E8}[87.3, 89.4] \\
      & Baseline           & 81.2                         & [76.4, 86.0]                         & 62.8                         & [60.9, 64.7]                         \\ \cline{2-6} 
      & \textit{p-values}  & \multicolumn{2}{l}{\textbf{Clustering: $<0.001$}}                   & \multicolumn{2}{l}{\textbf{Clustering: $<0.001$}}                   \\
      & \textit{}          & \multicolumn{2}{l}{\textbf{Iteration: $<0.001$}}                    & \multicolumn{2}{l}{\textbf{Iteration: $<0.001$}}                    \\ \hline
\end{tabular}
\end{table}
A summary of these results can be observed in Table \ref{ANOVA_RESULTS}, which reports the 95\% confidence intervals of each method for both the AUC and S percentages across the four iterations per process discovery algorithm, as well as the p-values of the clustering and iteration factors obtained with two-way analysis of variance (ANOVA). The results outline that simplicity values significantly depend on both the clustering approach and iterations, whereas the AUC is impacted by both with the HM algorithm.

Finally, Figure \ref{TIMING_ANALYSIS} shows the timing analysis of the four methods in terms of time required for batch processing of the event logs during online adaptation at each iteration for all the methods. As expected, since the baseline approach only involves applying the process discovery algorithm at each iteration with new data, its required time is negligible, while the other approaches also require computing diagnoses and applying clustering. In particular, such an increase is mostly due to the number of traces to check against the process models through conformance checking to extract diagnoses, which progressively worsens the required time at each iteration since new traces are added. However, the latency is still acceptable for batch processing scenarios.

\subsection{Discussion and limitations}
The results outlined the ability of the online adaptation framework to build specific and interpretable process models as new data are collected from the system. This approach allows for augmenting the knowledge base with new process models of adequate complexity and specificity, capturing the clinical pathways followed by patients affected by different diseases. Compared to the baseline approach, which updates a single process model by incorporating new data with the historical data, the proposal distributes knowledge among different process models adaptively by computing the differences between existing knowledge and new knowledge.

While the approach allows adaptive integration of new knowledge, the methodology and experimentation are affected by a few limitations:
\begin{itemize}
    \item{\textbf{Privacy concerns}: }While the incorporation of new patients' data is instrumental to understanding the healthcare processes and their bottlenecks, these data are highly sensitive and require ensuring privacy mechanisms to maintain confidentiality, such as employing $k$-anonimity \cite{rafiei2021groupbasedpppm}.
    \item{\textbf{Incremental discovery and repair}: }Although the approach allows clustering new traces and building ad-hoc process models, other paradigms such as incremental process discovery \cite{schuster2024analyzinghcprocesses} and process repair \cite{vinci2025balancing} can be integrated into the methodology as additional techniques for enhancing existing process models instead of creating new ones from the clustered traces.
    \item{\textbf{External validation}: }Whereas the Synthea dataset ensures both replicability and the ability to perform a thorough controlled experimentation, validation with actual, real-life electronic medical records would strengthen the external validity of the results.
\end{itemize}

Overall, despite these limitations, the proposed online adaptation framework offers a promising direction for developing flexible, data-driven representations of clinical pathways that evolve alongside incoming information. By supporting more granular and adaptive process models, the approach contributes to bridging the gap between static process discovery methods and the dynamic nature of real-world healthcare environments, laying the groundwork for future research on scalable, privacy-preserving, and clinically validated adaptive process mining solutions.

\section{Conclusions}
\label{sec:conclusions}

The differences in patients' disease development and treatment call for adequate evidence-based support to tailor the clinical pathway to the specific cases at hand. To support the identification and modeling of clinical pathways, the large amount of evidence collected from historical and ongoing clinical activities can be leveraged. Although process mining has been recently adopted in healthcare applications to study the evolution and management of patient treatments due to its process-based approach and interpretability, these methods lack a systematic method to adaptively identify clinical pathways and encode them into quality process models. In view of this, we proposed a two-phase process mining-based method to 1) adaptively identify clinical pathways and 2) high-quality modeling through discovery algorithms. We applied the method to the Synthea COVID-19 dataset, which establishes a realistic benchmark of a highly variable disease that may cause diverse complications and result in different outcomes. Results highlighted that our method enables expanding the knowledge base of clinical pathways with sufficient precision, striking a balance between specificity and generalization. Specifically, our method is able to maintain a high AUC ($\geq89.25$) across the four iterations, peaking at $94.82\%$ and only dropping down to 80.58\%, while the baseline has been shown to drop to as low as 67.04\%. The best result is achieved with the ILP combined with DBScan, achieving up to 95.62\% AUC while maintaining an arc-degree simplicity of 67.11\%.

Future work involves improving the adaptation cycle of the online phase of our method by reducing the number of models, which can become significantly higher as more iterations are performed. In addition, we plan on performing more experimentation with other datasets involving other diseases, and add an explanation layer to our method to investigate the root causes of clinical pathway deviations from the knowledge base.

\printbibliography

\end{document}